\documentclass{article}
\pdfoutput=1
\usepackage{spconf,amsmath,graphicx}
\usepackage{multirow}
\usepackage{enumitem}

\usepackage[american]{babel}
\usepackage{microtype}
\usepackage{times}
\usepackage{epsfig}
\usepackage{graphicx}
\usepackage{amsmath}
\usepackage{amssymb}
\usepackage{hyperref}
\usepackage{color}

\usepackage{bm}
\usepackage{amsmath}
\usepackage{algorithm}
\usepackage{algorithmic}
\usepackage{amssymb}
\usepackage{booktabs}
\usepackage{subfigure}
\usepackage{textcomp}
\usepackage{multirow}

\usepackage{comment}
\usepackage{amsthm}
\usepackage{caption}
\usepackage{arabtex}
\title{RD-NAS: Enhancing One-shot Supernet Ranking Ability via Ranking Distillation from Zero-cost Proxies}
\name{
\begin{tabular}{c}{}
Peijie Dong$^{1}$, Xin Niu$^{1,\ast}$\thanks{*Corresponding author}, Lujun Li$^{2}$, Zhiliang Tian$^{1}$, Xiaodong Wang$^{1}$ \\
Zimian Wei$^{1}$, Hengyue Pan$^{1}$, Dongsheng Li$^{1}$
\end{tabular}
}

\address{$^1$ School of Computer, National University of Defense Technology, Hunan, China\\
$^2$ Chinese Academy of Sciences, Beijing, China}

\begin{document}
% \ninept
\maketitle
\begin{abstract} Neural architecture search (NAS) has made tremendous progress in the automatic design of effective neural network structures but suffers from a heavy computational burden. One-shot NAS significantly alleviates the burden through weight sharing and improves computational efficiency. Zero-shot NAS further reduces the cost by predicting the performance of the network from its initial state, which conducts no training. Both methods aim to distinguish between "good" and "bad" architectures, i.e., ranking consistency of predicted and true performance. In this paper, we propose Ranking Distillation one-shot NAS (RD-NAS) to enhance ranking consistency, which utilizes zero-cost proxies as the cheap teacher and adopts the margin ranking loss to distill the ranking knowledge. Specifically, we propose a margin subnet sampler to distill the ranking knowledge from zero-shot NAS to one-shot NAS by introducing Group distance as margin. Our evaluation of the NAS-Bench-201 and ResNet-based search space demonstrates that RD-NAS achieve 10.7\% and 9.65\% improvements in ranking ability, respectively. Our codes are available at https://github.com/pprp/CVPR2022-NAS-competition-Track1-3th-solution
\end{abstract}


\section{Introduction}

\begin{figure}[ht]
%   \centering
  \includegraphics[width=0.9\linewidth]{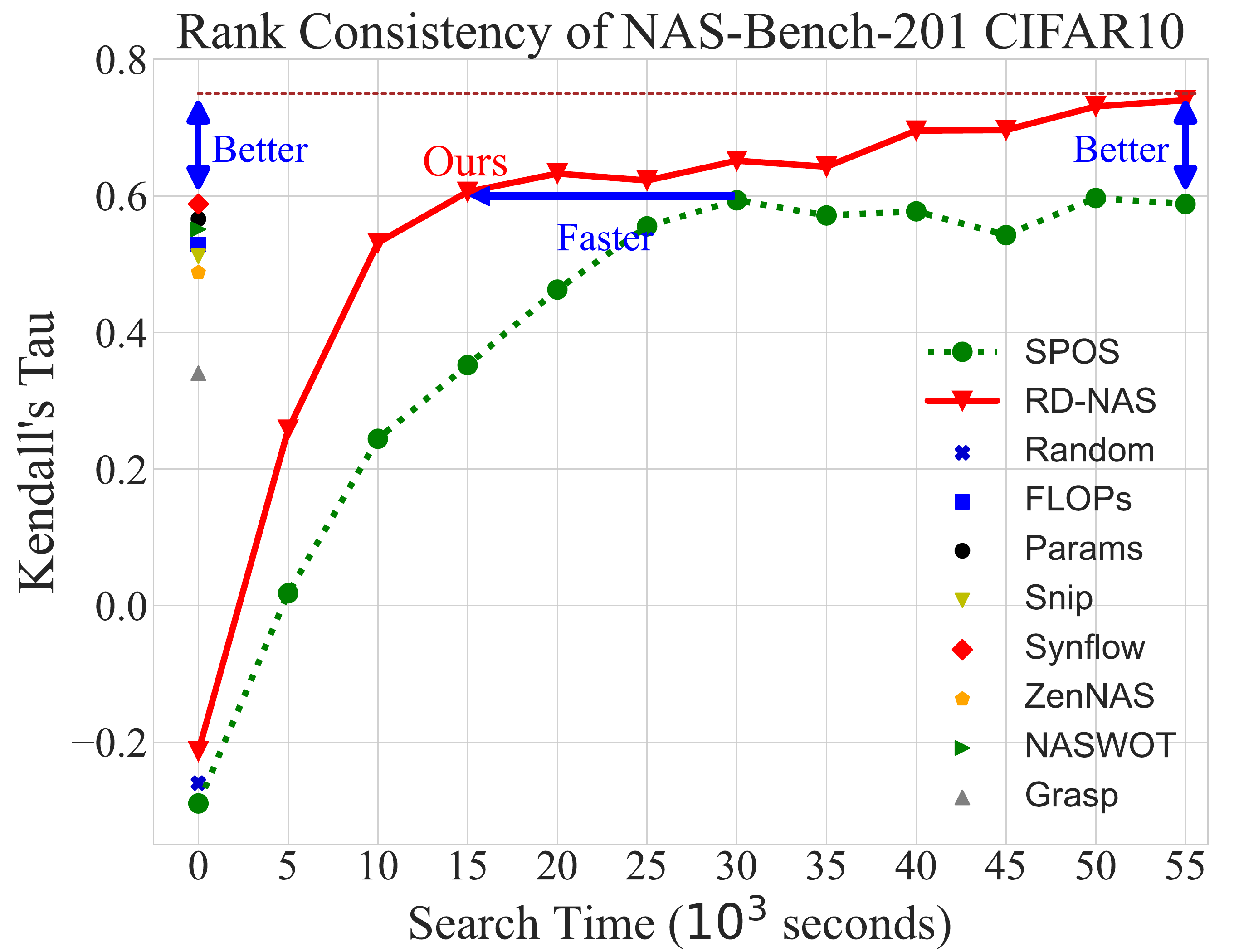}
  \vspace{-1em}
  \caption{The ranking consistency of one-shot NAS (SPOS~\cite{Guo2020SinglePO}) and zero-shot NAS(FLOPs~\cite{Abdelfattah2021ZeroCostPF}, Params~\cite{Abdelfattah2021ZeroCostPF}, Snip~\cite{Abdelfattah2021ZeroCostPF}, Synflow~\cite{Abdelfattah2021ZeroCostPF}, ZenNAS~\cite{Lin2021ZenNASAZ}, NASWOT~\cite{Mellor2021NeuralAS}, Grasp~\cite{Abdelfattah2021ZeroCostPF}). Our proposed RD-NAS achieves higher Kendall's Tau than both one-shot and zero-shot NAS and converges faster than the baseline ~\cite{Guo2020SinglePO}}\label{fig:motivation}
  \vspace{-2em}
\end{figure}

Neural Architecture Search (NAS) has sparked increased interest due to its remarkable progress in a variety of computer vision tasks~\cite{liu2020block,Lin2021ZenNASAZ,li2021nas,linas2,lichengp}. It aims to reduce the cost of human efforts in manually designing network architectures and discover promising models automatically. Early NAS works~\cite{zoph2017neural,real2019regularized,wei2022convformer} take thousands of GPU hours in search cost, and ENAS~\cite{pham2018efficient} first attempt at weight-sharing techniques to accelerate the searching process. The key to weight-sharing-based methods is an over-parameterized network, \textit{a.k.a.} supernet, that encompasses all candidate architectures in the search space. The weight-sharing-based NAS approaches~\cite{Guo2020SinglePO,liu2018darts} are called one-shot NAS since it only requires the cost of training one supernet. Despite the high efficiency of one-shot NAS, it is theoretically based on the ranking consistency assumption, \textit{a.k.a.} the estimated performance of candidate architectures in supernet should be highly correlated with the true performance of the corresponding architecture when trained from scratch. However, due to the nature of the weight-sharing approach, the subnet architectures interfere with each other, and the estimated accuracy from the supernet is inevitably averaged.

\begin{figure*}[ht]
  \vspace{-2em}
  \centering
  \includegraphics[width=0.85\textwidth]{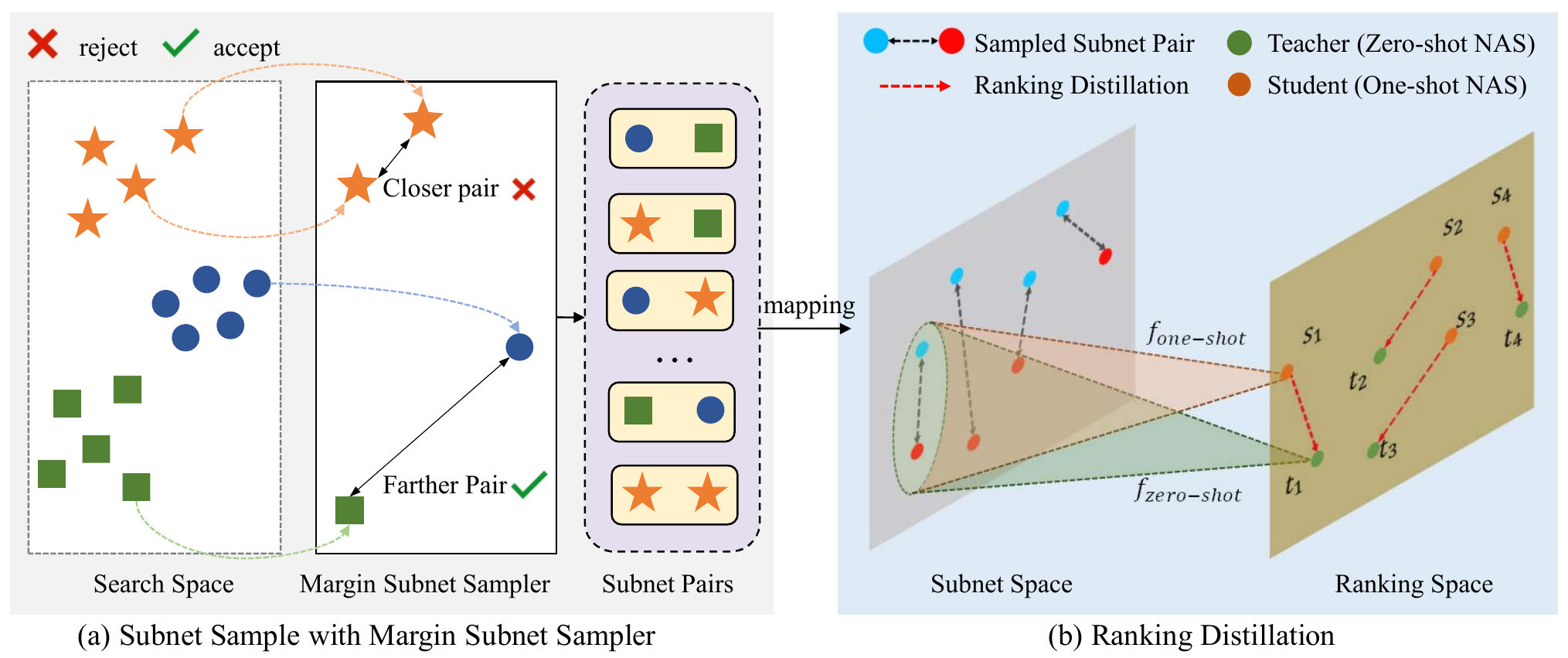}
  \caption{Overview of the proposed RD-NAS. (a) Margin Subnet Sampler is utilized to measure the distance of subnets in the search space and sample subnet pair with margin. (b) The subnet pairs sampled in the subnet space are mapped to the ranking space by the one-shot and Zero-shot methods, and the ranking knowledge from zero-shot NAS is transferred to the one-shot NAS via ranking distillation.}
  \label{fig:main}
  \vspace{-1.5em}
\end{figure*}

A recent trend of NAS focuses on zero-cost proxies~\cite{Mellor2021NeuralAS, Abdelfattah2021ZeroCostPF}, which aims to estimate the relative performance of candidate architecture from a few mini-batch of data without training the model. The approach of inferring the trained accuracy directly from the initial state of the model is called zero-shot NAS since it does not involve any training. However, there is a clear preference for zero-shot NAS that affects its ability to find good architectures~\cite{Ning2021EvaluatingEP}.

To alleviate the ranking disorder caused by weight-sharing, some predictor-based methods~\cite{Hu2021RankNASEN, Wang2021RANKNOSHEP} use the ranking loss to enhance the ranking ability of the model, but the challenge is that acquiring training samples (pairs of (model, accuracy)) is computationally expensive. To improve the ranking consistency economically, we resort to zero-shot NAS as an inexpensive teacher and measure the ranking relationship between a pair of subnets without employing a predictor. In recent years, the experimental investigations~\cite{Abdelfattah2021ZeroCostPF} enhanced the existing NAS algorithms using zero-cost proxies to initialize the search algorithm at the beginning of the search process, which accelerated the convergence speed of the supernet. Unlike its limited use, we incorporate zero-cost proxies into the whole training procedure with ranking knowledge distillation~\cite{li2022norm,li2021seg,li2022self,li2022SFF,lishadow,li2020explicit,li2022tf}.

We observe the complementarity between one-shot NAS and zero-shot NAS in Fig.\ref{fig:motivation}. Zero-cost proxies show good initial ranking consistency, but the upper-performance limit is not as high as one-shot NAS, while one-shot NAS has lower ranking performance in the early stage. This inspires us to bridge one-shot NAS and zero-shot NAS by utilizing zero-cost proxies as ranking teachers. In this paper, we propose a novel Ranking Distillation one-shot NAS (RD-NAS) to transfer the ranking knowledge from Zero-cost proxies to one-shot supernet. 

Our contributions are summarized as follows: 
(1) We propose a Ranking Distillation framework via margin ranking loss to distill knowledge from zero-cost proxies and improve the ranking ability of one-shot NAS. (2) We propose a margin subnet sampler to reduce the uncertainty caused by zero-cost proxies by introducing group distance. (3)  We demonstrate the effectiveness of RD-NAS on two types of benchmark, \textit{a.k.a.} ResNet-like search space, and NAS-Bench-201, which achieve 10.7\% and 9.65\% improvement on Spearman and Kendall's Tau, respectively.

\section{Ranking Distillation One-shot NAS}

Ranking Distillation one-shot NAS is a framework for distilling ranking knowledge from zero-shot NAS to improve the ranking ability of one-shot NAS. In Sec.\ref{preliminaries}, we introduce how to measure the effectiveness of zero-cost proxies in the context of ranking and introduce a new criteria. In Sec.\ref{rd}, we propose the margin subnet sampler with margin ranking loss to perform the ranking distillation.

\subsection{How to measure Zero-cost Proxies}\label{preliminaries}

Given a candidate architecture $\alpha_i\in \Omega, (i=1,2,...,N)$ in the search space $\Omega$, zero-cost proxies can quickly estimate the relative score $\pi^{z}_i\in \mathbb{R}$ of the $i$-th architecture, which can obtain the relative ranking relationship with other counterparts. However, there is a certain gap between the score $\pi^{z}_i$ obtained by zero-cost proxies and truth scores $\pi^{*}_i\in \mathbb{R}$ obtained by stand-alone training. We introduce $\sigma_i$ to estimate the variance of the zero-cost score of the $i$-th architecture, and the equation $\pi^{*}_i=\pi^{z}_i+\sigma_i$ holds. We expect to introduce truth ranking correlation between architectures into the supernet optimization process. To measure the estimation ability of zero-cost proxies, we incorporate the objective function: 
% {\setlength\abovedisplayskip{0.1cm}
% \setlength\belowdisplayskip{0.1cm}
\begin{align}
    max_{i,j} P(\pi^{*}_i>\pi^{*}_j) &= max_{i,j} P(\pi^{z}_i-\pi^{z}_j+\sigma_i-\sigma_j>0)\notag \\
    &\approx max_{i,j} P(\pi^{z}_i-\pi^{z}_j>0)
\end{align}\label{equation:obj}
The variance of a certain zero-cost proxy can be considered as a constant scalar, so that $\mathbb{E}[\sigma_i-\sigma_j]\approx 0$, where we use zero-cost score $\pi^{z}$ to estimate the ground truth $\pi^{*}$. Here we simplify Kendall's tau and introduce Concordant Pair Ratio (CPR) $\delta$ to evaluate the pair-wise ranking consistency. We define the subnet pairs that satisfying $(\pi^*_i-\pi^*_j)(\pi^z_i-\pi^z_j) > 0$ as the concordant pair. The Concordant Pair Ratio coefficient $\delta$ is defined as:
% {\setlength\abovedisplayskip{0.1cm}
% \setlength\belowdisplayskip{0.1cm}
\begin{equation}
    \delta = \frac{\sum_{i,j}1((\pi^*_i-\pi^*_j)(\pi^z_i-\pi^z_j)>0)}{\sum_{i,j}1}
\end{equation}
\noindent which is in the range $0\leq \delta \leq 1$. The closer $\delta$ is to 1, the better estimation of the zero-cost proxy for ground truth and \textit{vice versa}. 

\subsection{Ranking Distillation}\label{rd}

To transfer the ranking knowledge obtained from zero-cost proxies to one-shot supernet, we propose a ranking distillation to further improve the ranking consistency of the one-shot NAS. In Sec.\ref{pair}, we introduce the margin ranking loss to distill the ranking knowledge from zero-cost proxies, but the margin is hard to set in one-shot training. Therefore, we propose a margin subnet sampler to sample subnet pairs from the subnet space in Sec.\ref{margin}.

\subsubsection{Margin Ranking Loss}\label{pair}

Margin ranking loss is adopted to constrain the optimization of the supernet from the perspective of ranking distillation. Unlike the standard pairwise ranking loss~\cite{Yu2021LandmarkRR}, we do not have the true rank, only the estimated rank provided by zero-cost proxies, which suffer from uncertainty. Margin ranking loss can handle difficult subnet pairs by introducing a fixed penalty margin $m$. The margin ranking loss of random-sampled subnet pair ($\alpha_i$, $\alpha_j$) is as follows:
\begin{equation}
    \resizebox{0.91\hsize}{!}{
    $\mathcal{L}^{rd}_{(\alpha_i,\alpha_j)}\left(\theta^{s}\right)=\max \left[0, m - \left(\mathcal{L}^{ce}\left(x, \theta_{\alpha_{i}}^{s}\right)-\mathcal{L}^{ce}\left(x, \theta_{\alpha_{j}}^{s}\right)\right)\right]$
    }
\end{equation}
\noindent where $\theta^s$ denotes the supernet in one-shot NAS, and $\mathcal{L}^{ce}(\cdot)$ denotes the cross-entropy loss. However, such a fixed learning objective is not realistic as the training goes with inconsistent variation, which might limit the discriminative training of supernet. Instead of using a fixed margin $m$, we proposed a margin subnet sampler to control the margin in the subnet sampling process.

% The training loss is $\mathcal{L} = \mathcal{L}^{ce} + \lambda \mathcal{L}^{rd}$.

\begin{table}[]
\caption{The comparison of Kendall's Tau w.r.t different methods on ResNet-like search space. "MSS" denotes the margin subnet sampler.}
\vspace{-3mm}
\centering
\resizebox{0.9\linewidth}{!}{
    \begin{tabular}{c|lc}
    \hline
    \multicolumn{1}{l|}{Type}      & Method         & Spearman $r$(\%)      \\ \hline
    \multirow{3}{*}{Zero-cost Proxies} & Params~\cite{Abdelfattah2021ZeroCostPF}         & 67.23 \\
                              & FLOPs~\cite{Abdelfattah2021ZeroCostPF}          & 78.43 \\
                              & ZenNAS~\cite{Lin2021ZenNASAZ} & 81.27 \\ \hline
    \multirow{5}{*}{One-shot NAS}  & SPOS~\cite{Guo2020SinglePO} & 72.49 \\
                            %   & FairNAS~\cite{chu2019fairnas} & 78.20 \\
                              & Sandwich~\cite{Yu2019AutoSlimTO} & 74.86 \\
                              & RD-NAS w/o MSS & 81.14 \\
                              & \textbf{RD-NAS w/ MSS}  & \textbf{83.20} \\ \hline
    \end{tabular}
}\label{tab:resnet-like}
  \vspace{-1em}
\end{table}

\subsubsection{Margin Subnet Sampler}\label{margin}

It is difficult to tune using a fixed margin, and the margin changes as training progresses. The effectiveness of margin ranking loss is greatly affected by the subnet pair chosen. To tackle the problems, we propose a margin subnet sampler, which affects the margin by controlling the sampled subnets. Intuitively, we expect the sampled subnet pairs to cover the entire search space, which can improve the generalization ability. We first randomly select $n$ pairs of subnet $\{(\alpha_i,\alpha_j)\}_n$ from the search space, then we calculate the distance of the subnet pair with distance function $f(\cdot)$ mapping from subnet space $\mathbb{R}^{+}$ to ranking space $\mathbb{R}$ and filter out the $c$ pairs of subnet $\{(\alpha_i,\alpha_j)\}_c = \{(\alpha_i, \alpha_j) | f(\alpha_i, \alpha_j) > \mu\}$, where distances less than the threshold $\mu$ is selected. 

Concerning the subnet distance measure $f(\cdot)$, by encoding the subnet as a sequence composed of candidate choices, the Hamming distance $f_{h}(\cdot)$ is computed as the number of elements that are different in the two subnet sequences. However, Hamming distance treats all candidate operations as equal and ignores the variability between them. To make the distance measure more discriminative, we propose the Margin Subnet Sampler (MSS) with Group distance $f_{g}(\cdot)$, which divides the candidate operations into groups with and without parameters. The intra-group distance is set small, and the inter-group distance is set large.
{\setlength\abovedisplayskip{0.1cm}
\setlength\belowdisplayskip{0.1cm}
\begin{equation}
f_g^k(\alpha_i^k, \alpha_j^k)=\left\{
\begin{array}{rcl}
2 f_h(\alpha_i^{k}, \alpha_j^{k}),   &      & \beta(\alpha_i^k, \alpha_j^k)=0\\
0.5 f_h(\alpha_i^{k}, \alpha_j^{k}),     &      & \beta(\alpha_i^k, \alpha_j^k)=1
\end{array} \right.
\end{equation}}
\noindent where $\beta(\cdot)$ is an indicator function of whether $(\alpha_i^k, \alpha_j^k)$ are intra-group and $f_g^k(\cdot)$ denotes the group distance of the $k$-th operation of the subnet encoding.

\begin{table}[]
    \centering
    \small 
    \caption{Ranking correlation on NAS-Bench-201 CIFAR-10. $\tau$, r, $\rho$, and $\delta$ refers to Kendall's Tau, Pearson, Spearman and CPR, respectively. "Random" refers to the rank correlation on random initialized model.}
    \vspace{-3mm}
    \centering
    \begin{tabular}{lcccc}
        \hline
                                                & $\tau$(\%)          & $r$(\%)             & $\rho$(\%)          & $\delta$(\%)  \\ \hline
        FLOPs~\cite{Abdelfattah2021ZeroCostPF}                                   & 52.92          & 47.12          & 72.44          & 78.49 \\
        Params~\cite{Abdelfattah2021ZeroCostPF}                                  & 56.64          & 49.14          & 74.70          & 77.49 \\
        Snip~\cite{Abdelfattah2021ZeroCostPF}    & 51.16          & 28.20          & 69.12          & 80.49  \\
        Synflow~\cite{Abdelfattah2021ZeroCostPF} & 58.83          & 44.90          & 77.64          & 84.99 \\
        ZenNAS~\cite{Lin2021ZenNASAZ}            & 48.80          & 55.72          & 67.83          & 44.99 \\
        NASWOT~\cite{Mellor2021NeuralAS}           & 55.13        & 86.71          & 74.93          & 70.99 \\ \hline 
        Random                                & -26.32         & -21.43         & -22.48         & 1.99    \\ 
        SPOS~\cite{Guo2020SinglePO}         & 77.43          & 89.55          & 92.73          &  100.0   \\
        FairNAS~\cite{chu2019fairnas}     & 70.62          & 82.81          & 88.74          & 100.0   \\ 
        \textbf{RD-NAS(Ours)} & \textbf{87.08} & \textbf{95.00} & \textbf{97.34} & \textbf{100.0} \\
        \hline
    \end{tabular}\label{one-zero-rank}
    % \tabnote{$^{\rm \dag}$ Methods that need One-shot training.}
\vspace{-3mm}
\end{table}

\begin{table*}
    \centering
    \small
    \caption{Searching results on NAS-Bench-201 dataset~\cite{nas201}. Our method obtains 1.12\%, 3.76\%, and 4.18\% absolute accuracy improvement compared with SPOS on the test sets of CIFAR-10, CIFAR-100, and ImageNet-16-120, respectively. Note that results for other weight-sharing methods are reported by NAS-Bench-201 benchmark~\cite{nas201}, which only uses the labels of the dataset as the true performance of the searched architecture for a fair comparison.}
    \label{tab:201-sota}%
    \vspace{-3mm}
    % \resizebox{170mm}{!}{
    \begin{tabular}{lllllll}
        \toprule
        \multirow{2}[4]{*}{Method} & \multicolumn{2}{c}{CIFAR-10 Acc. (\%)} & \multicolumn{2}{c}{CIFAR-100 Acc. (\%)} & \multicolumn{2}{c}{ImageNet-16-120 Acc. (\%)}                                                     \\
        \cmidrule{2-7}             & Validation                             & Test                                    & Validation                                    & Test            & Validation     & Test           \\
        \midrule
        %RSPS~\cite{RSA}            & 84.16$\pm$1.69& 87.66$\pm$1.69& 59.00$\pm$4.60& 58.33$\pm$4.34  & 31.56$\pm$3.28 & 31.14$\pm$3.88 \\
        GDAS~\cite{GDAS}           & 90.00$\pm$0.21& 93.51$\pm$0.13& \textbf{71.14$\pm$0.27}& 70.61$\pm$0.26  & 41.70$\pm$1.26 & 41.84$\pm$0.90 \\
        %SETN~\cite{SETN}           & 84.04$\pm$0.28& 87.64$\pm$0.00& 58.86$\pm$0.04& 59.05$\pm$0.20  & 33.06$\pm$0.02 & 32.51$\pm$0.21 \\
        DSNAS~\cite{DSNAS}         & 89.66$\pm$0.29& 93.08$\pm$0.13& 30.87$\pm$16.40& 31.01$\pm$16.38 & 40.61$\pm$0.09 & 41.07$\pm$0.09 \\
        PC-Darts~\cite{pcdarts}    & 89.96$\pm$0.15& \textbf{93.41$\pm$0.30}& 67.12$\pm$0.39& 67.48$\pm$0.89  & 40.83$\pm$0.08 & 41.31$\pm$0.22 \\
        \midrule 
        NASWOT~\cite{Mellor2021NeuralAS} & 89.14$\pm$1.14 & 92.44$\pm$1.13 & 68.50$\pm$2.03 & 68.62$\pm$2.04 & 41.09$\pm$3.97 & 41.31$\pm$4.11 \\
        EPENAS~\cite{Lopes2021EPENASEP} & 89.90$\pm$0.21 & 92.63$\pm$0.32 & 69.78$\pm$2.44 & 70.10$\pm$1.71 & 41.73$\pm$3.60 & 41.92$\pm$4.25 \\
        \midrule
        Random            & 83.20$\pm$13.28 & 86.61$\pm$13.46 & 60.70$\pm$12.55 & 60.83$\pm$12.58 & 33.34$\pm$9.39 & 33.13$\pm$9.66 \\
        SPOS~\cite{Guo2020SinglePO}  & 88.40$\pm$1.07& 92.24$\pm$1.16& 67.84$\pm$2.00& 68.07$\pm$2.25  & 39.28$\pm$3.00 & 40.28$\pm$3.00 \\
        % FairNAS~\cite{chu2019fairnas} & 90.07$\pm$0.57 & 93.23$\pm$0.18 & 70.94$\pm$0.94 & 71.00$\pm$1.46 & 41.90$\pm$1.00 & 42.19$\pm$0.31 \\
        \textbf{RD-NAS(Ours)}      & \textbf{90.44$\pm$0.27}& 93.36$\pm$0.04& 70.96$\pm$2.12& \textbf{71.83$\pm$1.33}      & \textbf{43.81$\pm$0.09} & \textbf{44.46$\pm$1.58}     \\
        \midrule
        Optimal                    & 91.61& 94.37& 73.49& 73.51& 46.77& 47.31\\
        \bottomrule
    \end{tabular}
    %}
    \vspace{-3mm}
\end{table*}

\section{Experiments}

The experiments in Sec.\ref{exp:resnet-like} and Sec.\ref{exp:nasbench201} verify the effectiveness of the RD-NAS on the ResNet-like search space and NAS-Bench-201. Then, we conduct the ablation study for RD-NAS.

\subsection{Searching on ResNet-like Search Space}\label{exp:resnet-like}

\noindent\textbf{Setup.} Presented in CVPR NAS Workshop, the ResNet-like search space is based on ResNet48 on the ImageNet~\cite{russakovsky2015imagenet}, whose depth and expansion ratio are searchable. Specifically, the search space consists of four stages with different depths \{5,5,8,5\}, while the candidate expansion ratio is ranging from 0.7 to 1.0 at 0.05 intervals, and the channel number should be divided by 8. The search space size is $5.06 \times 10^{19}$ in total. Our proposed RD-NAS won the third place at CVPR 2022 NAS Track1. We conduct a comprehensive comparison of the zero-shot~\cite{Lin2021ZenNASAZ} and one-shot methods~\cite{Guo2020SinglePO,chu2019fairnas,Yu2019AutoSlimTO}, and the Spearman's ranking correlation is adopted to measure the ranking ability. We first pre-train the supernet for 90 epochs in ImageNet and then train the supernet with a learning rate of 0.001 with a batch size of 256 for 70 epochs. The SGD optimizer is adopted with a momentum of 0.9. 

\noindent\textbf{Results.} The results of Spearman's ranking correlation are shown in Tab.\ref{tab:resnet-like}. RD-NAS could achieve significantly higher ranking correlations than both the zero-shot and one-shot NAS, especially using the margin subnet sampler. The Kendall's Tau of RD-NAS surpasses the baseline method~\cite{Guo2020SinglePO} by 10.71\% and surpasses the zero-cost proxy - "FLOPs", demonstrating that using zero-cost proxies as teachers for distillation can effectively mitigate the ranking disorder problem.

\subsection{Searching on NAS-Bench-201}\label{exp:nasbench201}

\noindent\textbf{Setup.} NAS-Bench-201~\cite{nas201} provides the performance of 15,625 subnets on CIFAR-10, CIFAR-100, and ImageNet-16-120. In our experiments, the training settings are consistent with NAS-Bench-201. Following SPOS~\cite{Guo2020SinglePO}, we formulate the search space of NAS-Bench-201~\cite{nas201} as a single-path supernet.  The distance threshold $\mu$ is set to 6.7.

\noindent\textbf{Evaluation Criteria.} The correlation criteria used in NAS-Bench-201 are Pearson $r$, Kendall' Tau $\tau$~\cite{Sen1968EstimatesOT}, Spearman $\rho$, and CPR $\delta$. Pearson measures the linear relationship between the variables, while Kendall’s Tau, Spearman, and CPR measure the monotonic relationship. The ranking correlations are computed using 200 randomly selected subnets. 

\noindent\textbf{Results.} Our method obtains 2.04\%, 3.12\%, and 4.53\% absolute accuracy improvement on the validation sets of CIFAR-10, CIFAR-100, and ImageNet-16-120 in Tab.\ref{tab:201-sota}, compared with SPOS, which further certifies that RD-NAS enables to boost the ranking ability of one-shot NAS. Even compared with the recent NAS methods (e.g., PC-Darts~\cite{pcdarts} and GDAS~\cite{GDAS}), our method achieve comparable performance on the NAS-Bench-201. Especially for ImageNet-16-120, our RD-NAS achieves state-of-the-art performance that surpasses all other methods. The results on Tab.\ref{one-zero-rank} reveal that our RD-NAS achieve better ranking correlation than one-shot and zero-shot NAS.

\subsection{Ablation Study}

The ablation study of the distance measurement and training strategy on NAS-Bench-201 CIFAR-10 is shown in Tab.\ref{tab:ablation}. The one-shot training with uniform sampling strategy\cite{Guo2020SinglePO} is employed as the "Random" baseline. The results of Ranking Distillation exhibit the effectiveness of margin ranking loss. Moreover, using hamming or adaptive distance to sample subnets can further improve the rank consistency. The combination of margin subnet sampler and ranking distillation improves the Kendall's Tau from 77\% to 87\%.

% \vspace{-3mm}

\begin{table}[]
	\caption{Ablation study of RD-NAS on NAS-Bench-201 CIFAR-10. "Random" denotes the baseline SPOS\cite{Guo2020SinglePO}, "Margin" and "Hamming" denotes margin subnet sampler with Group and Hamming distance, respectively. "RD" denotes Ranking Distillation.}
    \vspace{-3mm}
    \small 
	\centering
	\begin{tabular}{cccc|c}
		\hline
		Random       & Margin    & Hamming    & RD        & Kendall' Tau(\%) \\ \hline
		$\checkmark$ & -            & -            & -                    & 77.43      \\
		-            & -            & -            & $\checkmark$         & 79.37      \\
		-            & -            & $\checkmark$ & $\checkmark$         & 84.18      \\
		-            & $\checkmark$ & -            & $\checkmark$         & \textbf{87.08}      \\ \hline
	\end{tabular}\label{tab:ablation}
\vspace{-3mm}
\end{table}

\section{Conclusion}

In this paper, we present a novel ranking knowledge distillation for one-shot NAS, which we call RD-NAS, to mitigate the rank disorder problem. Specifically, our RD-NAS employs zero-cost proxies as teachers and transfers the ranking knowledge to one-shot supernet by margin ranking loss. To reduce the uncertainty of zero-cost proxies, we introduce the margin subnet sampler to increase the reliability of ranking knowledge. We conduct detailed experiments on ResNet-like search space and NAS-Bench-201 to demonstrate the effectiveness of RD-NAS. It is hoped that our work will inspire further research to exploit the strengths of zero-shot NAS.

%\vfill\pagebreak

\bibliographystyle{IEEEbib}
\bibliography{RDNAS}

\end{document}